# Inferring Shallow-Transfer Machine Translation Rules from Small Parallel Corpora

**Felipe Sánchez-Martínez**                          FSANCHEZ@DLSI.UA.ES
**Mikel L. Forcada**                                 MLF@DLSI.UA.ES
*Departament de Llenguatges i Sistemes Informàtics*
*Universitat d'Alacant, E-03071 Alacant (Spain)*

## Abstract

This paper describes a method for the automatic inference of structural transfer rules to be used in a shallow-transfer machine translation (MT) system from small parallel corpora. The structural transfer rules are based on alignment templates, like those used in statistical MT. Alignment templates are extracted from sentence-aligned parallel corpora and extended with a set of restrictions which are derived from the bilingual dictionary of the MT system and control their application as transfer rules. The experiments conducted using three different language pairs in the free/open-source MT platform Apertium show that translation quality is improved as compared to word-for-word translation (when no transfer rules are used), and that the resulting translation quality is close to that obtained using hand-coded transfer rules. The method we present is entirely unsupervised and benefits from information in the rest of modules of the MT system in which the inferred rules are applied.

## 1. Introduction

*Machine translation* (MT) may be defined as the use of a computer to translate a text from one natural language, the *source language* (SL), into another, the *target language* (TL). MT is difficult mainly because natural languages are highly ambiguous and also because two languages do not always express the same content in the same way (Arnold, 2003).

The different ways in which the MT problem has been approached may be classified according to the nature of the knowledge used in the development of the MT system. From this point of view, one can distinguish between corpus-based and rule-based approaches; although, hybrid approaches are also possible.

Corpus-based approaches to MT, such as *example-based* MT (EBMT; Nagao, 1984; Carl & Way, 2003) and *statistical* MT (SMT; Brown et al., 1993; Knight, 1999), use large collections of parallel texts as the source of knowledge from which the engine learns how to perform translations. A *parallel text* is a text in one language together with its translation into another language; a large collection of parallel texts is usually referred as a *parallel corpus*. Although corpus-based approaches to MT have grown in interest over the last years, they require large amounts, in the order of tens of millions of words, of parallel text to achieve reasonable translation quality (Och, 2005). Such a vast amount of parallel corpora is not available for many under-resourced language pairs demanding MT services.

Rule-based MT (RBMT) systems use knowledge in the form of rules explicitly coded by human experts that attempt to codify the translation process. RBMT systems heavily depend on linguistic knowledge, such as morphological and bilingual dictionaries (containing lexical, syntactic and even semantic information), part-of-speech disambiguation rules or manually disambiguated corpora, and a large set of rules. The process of building a RBMT system involves considerable human effort in order to develop the necessary linguistic resources (Arnold, 2003).

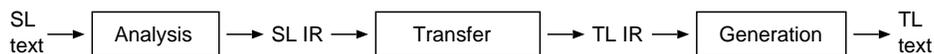

**Figure 1:** Scheme of a general transfer-based MT system.

Generally, RBMT systems work by parsing (or *analyzing*) the SL text, usually creating an intermediate (symbolic) representation (IR), from which the text in the TL is generated (Hutchins & Somers, 1992). According to the nature of the IR used, an RBMT system may be said to be either interlingua or transfer-based. An *interlingua* MT system uses a single IR that is independent of the languages involved in the translation; the advantage of using a language-independent IR is that no transfer module needs to be developed for each new language pair; as a disadvantage such an IR used is difficult to design and hard to implement, even more so, for open-domain tasks. In contrast, a *transfer-based* MT system uses two IRs, one for each of the languages involved; this has the advantage of easing the design and development of the IRs used, but at the cost of having to develop a transfer module for each new language pair.

*Transfer-based* MT systems usually work by applying, in addition to *lexical transfer* mappings, a set of *structural transfer* rules to the SL IR created during the analysis, in order to transform it into the TL IR from which the TL text is finally generated (see Figure 1). The level of analysis, and therefore the degree of abstraction provided by the IR, varies depending on how related the languages involved are. Translating between "distant" languages (such as English and Japanese) requires deep analysis (syntactic and semantic), while the translation between related languages (for example between Romance languages) can be achieved with shallow parsing. We will call this last type of transfer-based systems *shallow-transfer* MT systems.

## 1.1 Overview

This paper focuses on the automatic inference from small parallel corpora of the set of structural (shallow-)transfer rules that are used by shallow-transfer RBMT systems to convert a SL IR into the TL IR from which the TL text is generated. The development of such transfer rules requires qualified people to code them manually; therefore, their automatic inference may save part of this human effort. The method we present is entirely unsupervised and benefits from information in the rest of modules of the MT system in which the inferred rules are applied, in line with the method proposed by Sánchez-Martínez et al. (2008) to train part-of-speech taggers in an unsupervised way for their use in MT.

In our approach an existing bilingual dictionary is used to guide the inference of structural transfer rules (see below), and bilingual entries for that dictionary are not learned. This is because our approach is aimed at the inference of transfer rules from small parallel corpora[1] for their application in open-domain tasks. Note that small parallel corpora may be insufficient to obtain wide-coverage bilingual dictionaries, as demonstrated by the results obtained when translating through a state-of-the-art SMT system trained on the same small parallel corpora (see section 5). Notice that manually building a bilingual dictionary for a language pair is usually much easier than developing shallow structural transfer rules for it, moreover, the former task can be partially automated.

---

1. Small compared to the size of corpora commonly used to build corpus-based MT systems (Och, 2005).

The method we propose for the automatic inference of shallow-transfer rules from parallel corpora is based on the alignment template (AT) approach initially proposed for its use in the SMT framework (Och, 2002; Och & Ney, 2004). An AT can be defined as a generalization performed over aligned phrase[2] pairs (or *translation units*) by using word classes.

To adapt the AT approach to the RBMT framework, ATs are extended with a set of restrictions that control their application as structural shallow-transfer rules. To that end:

- the bilingual dictionary of the RBMT system in which the inferred rules will be integrated is used to ensure that the lexical content of each bilingual phrase pair extracted from the training corpus (see section 2.2) can be reproduced by the MT system;

- linguistically motivated word classes are used to generalize the extracted bilingual phrase pairs, deriving ATs from them; and,

- a set of restrictions, derived from the bilingual dictionary of the RBMT system, is attached to each AT to control its application as part of a transfer rule; this extension of the definition of AT will be called *extended AT*.

Once these extended ATs have been extracted from the training corpora, transfer rules are generated from them. In the experiments reported in section 5, shallow-transfer rules to be used by the Apertium MT engine (see appendix A) are generated directly in Apertium's XML-based structural transfer language. An interesting property of the inferred rules is that they are human-readable and may, therefore, be edited by human experts to improve their performance or supplemented with new rules; MT developers can use this method to infer an initial set of rules and then improve them by focusing on the more difficult issues.

Both this method (Sánchez-Martínez & Forcada, 2007) and its predecessor (Sánchez-Martínez & Ney, 2006) have already been presented in conferences; here, we explain the method in more detail, test it on two additional language pairs and use training corpora of different sizes so as to evaluate the impact of size on translation quality. Moreover, in this paper we perform a more detailed analysis of the inferred rules and the results obtained; to that end, we provide confidence intervals, which allow for a better interpretation of the results achieved. We will also discuss the process followed to build the parallel corpora used to learn transfer rules.

---

2. For the purpose of this paper, and to stick to the terminology used by Och and Ney (2004) in the definition of AT and by most SMT practitioners, by *phrase* we refer to any text segment, not necessarily a well-formed syntactic constituent.

## 1.2 Related Work

There have been other attempts to learn automatically or semi-automatically the structural transformations needed to produce correct translations into the TL. Those approaches can be classified according to the translation framework to which the learned rules are applied.

Some approaches learn transfer rules to be used in RBMT. Probst et al. (2002) and Lavie et al. (2004) developed a method to learn transfer rules for MT involving under-resourced languages (such as Quechua) with very limited resources. To this end, a small parallel corpus (of a few thousand sentences) is built with the help of a small set of bilingual speakers of the two languages. The parallel corpus is obtained by translating a controlled corpus from the language with more resources (English or Spanish) into the under-resourced language by means of an elicitation tool. This tool is also used to graphically annotate the word alignments between the two sentences. Finally, hierarchical syntactic rules, which can be seen as constituting a context-free transfer grammar, are inferred from the aligned parallel corpus.

Menezes and Richardson (2001) propose a method to infer transfer mappings (rules) between source and target languages. Prior to the acquisition of the transfer mappings, they align the nodes of the source and target parse trees by using an existing bilingual lexicon in which they look for word correspondences. Then, following a best-first strategy and using an small alignment grammar their method aligns the remaining (not-aligned) nodes. Once the alignments between the nodes of both parse trees have been obtained, frequencies are computed and sufficient context is retained to disambiguate between competing mappings at translation time. Our approach greatly differs from the one by Menezes and Richardson: (i) because they use a syntactic parser, a bilingual dictionary and an alignment grammar to obtain the word alignments from the sentence-aligned parallel corpus, while we only use statistical methods; (ii) because of how they use the bilingual dictionary, we use it to discard useless bilingual phrases and to derive restrictions to control the application of ATs, not for the computation of the word alignments; and (iii) because in our approach there is no ambiguity to solve at translation time.

Caseli et al. (2006) propose a method to infer bilingual resources (structural transfer rules and bilingual dictionaries) to be used in shallow-transfer MT from aligned parallel corpora. Previously to the generation of transfer rules, *alignment blocks* (sequences of aligned words) are built from the translation examples found in the parallel corpus by considering three different types of word alignments according to their geometry (crossings, unaligned words, etc.). Then, shallow-transfer rules are built in a three-step procedure. In the first step, they identify the patterns in two phases, monolingual and bilingual; then in a second step their method generates shallow-transfer rules by deriving monolingual and bilingual constraints, that can also be seen as the rule itself; finally, in a third step the rules are filtered in order to solve the ambiguity caused by rules matching the same SL sequence of words. The inferred rules are human-readable, as are those inferred with the method we propose, and may therefore be also edited by human experts. Our approach differs from that of Caseli et al. in how rules are induced: while our approach uses bilingual phrase pairs without being concerned about the type of alignments between the words, the way in which Caseli et al. induce rules depends on the type of the alignment blocks. In addition, our approach does not ever produce more than one rule matching the same sequence of SL items, and therefore no ambiguity needs to be solved. Furthermore, we do not infer a bilingual dictionary; instead, we use an existing bilingual dictionary to guide the inference of shallow-transfer rules, and to control the application of the inferred rules.

In the EBMT framework, some researchers have dealt with the problem of inferring a kind of translation rules called translation templates (Kaji et al., 1992; Brown, 1999; Cicekli & Güvenir, 2001). A *translation template* can be defined as a bilingual pair of sentences in which corresponding units (words or phrases) are coupled and replaced by variables. Liu and Zong (2004) provide an interesting review of the different research works dealing with translation templates. Brown (1999) uses a parallel corpus and some linguistic knowledge in the form of equivalence classes (both syntactic and semantic) to perform a generalization over the bilingual examples collected. The method works by replacing each word by its corresponding equivalence class and then using a set of grammar rules to replace patterns of words and tokens by more general tokens. Cicekli and Güvenir formulate the acquisition of translation templates as a machine learning problem, in which the translation templates are learned from the differences and similarities observed in a set of different translation examples, using no morphological information at all. Kaji et al. use a bilingual dictionary and a syntactic parser to determine the correspondences between translation units while learning the translation templates. Our approach differs from those applied in the EBMT framework because, on the one hand, the transfer rules generated through the method we propose are mainly based on *lexical forms* (consisting of lemma, lexical category and morphological inflection information) and, on the other hand, because they are flatter, less structured and non-hierarchical, which makes them suitable for shallow-transfer MT. Moreover, the way in which translation rules are chosen for application greatly differs from how they are chosen in the EBMT framework.

Finally, in the SMT framework the use of AT (Och & Ney, 2004) can be seen as an integration of translation rules into statistical translation models, since an AT is a generalization or an abstraction, of the transformations to apply when translating SL into TL by using word classes.

The rest of the paper is organized as follows: the next section reviews the alignment template (AT) approach; section 3 explains how ATs are extended with a set of restrictions in order to use them to generate shallow-transfer rules to be used in RBMT (section 4). Section 5 describes the experiments conducted and the results achieved. Finally, section 6 discusses the method described and outlines future research lines.

## 2. The Alignment Template Approach

The alignment template (AT) approach (Och, 2002; Och & Ney, 2004) was introduced in the SMT framework as one of the feature functions in the maximum entropy model (Och & Ney, 2002) to try to generalize the knowledge learned for a specific phrase to similar phrases.

An AT performs a generalization over bilingual phrase pairs using word classes instead of words. An AT $z = (S_m, T_n, A)$ consists of a sequence $S_m$ of $m$ SL word classes, a sequence $T_n$ of $n$ TL word classes, and a set of pairs $A = \{(i, j) : i \in [1, n] \land j \in [1, m]\}$ with the alignment information between the TL and SL word classes in the two sequences.

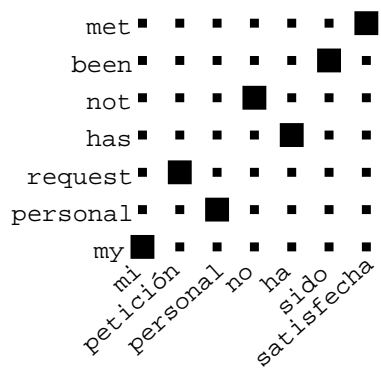

**Figure 2:** Alignment between the words in the English sentence *my personal request has not been met* and those in the Spanish sentence *mi petición personal no ha sido satisfecha*. The alignment information is represented as a binary matrix.

Learning a set of ATs from a sentence-aligned parallel corpus consists of: (i) the computation of the word alignments, (ii) the extraction of bilingual phrase pairs, and (iii) the generalization of such bilingual phrase pairs by using word classes instead of the words themselves.

## 2.1 Word Alignments

A variety of methods, statistical (Och & Ney, 2003) or hybrid (Caseli et al., 2005),[3] may be used to compute word alignments from a (sentence-aligned) parallel corpus. In the experiments reported in section 5, word alignments are obtained by training classical statistical translation models to translate from language $L_1$ to language $L_2$ (and vice versa) and then computing the Viterbi alignments under the previously estimated translation models. The Viterbi alignment between SL and TL sentences is defined as the alignment whose probability is maximal under the translation models previously estimated. The resulting Viterbi alignments $A_1$ and $A_2$ (one for each translation direction) are *symmetrized* through the refined intersection method proposed by Och and Ney (2003, p. 33). *Symmetrization* is needed in order to allow a SL word to be aligned with more than one TL word; otherwise, wrong alignments are obtained when a SL word actually corresponds to more than one TL word.

Figure 2 shows the word alignment in a Spanish–English sentence pair. The alignment information is represented as a binary matrix in which a value of 1 (large black squares) means that the words at the corresponding positions are aligned; analogously, a value of 0 (small black squares) means that the words are not aligned.

---

3. Caseli et al.'s (2005) method is hybrid because prior to the application of heuristics, it uses a statistical tool (NATools) to obtain a probabilistic bilingual dictionary (Simões & Almeida, 2003).

### 2.1.1 Training

In order to train the translation models and to calculate the Viterbi alignments of each pair of aligned sentences found in the training corpus the free/open-source `GIZA++` toolkit[4] (Och & Ney, 2003) is used with default parameters.

The computation of the word alignments consists of:

1. training the IBM model 1 (Brown et al., 1993) for 5 iterations; in this model, word order does not affect the alignment probabilities;

2. training the HMM alignment model (Vogel et al., 1996) for 5 iterations; this alignment model has the property of making alignment probabilities explicitly dependent on the alignment position of the previous word;

3. training the IBM model 3 (Brown et al., 1993) for 5 iterations; in this model, the probability of an alignment depends on the positions of the aligned words and on the length of SL and TL sentences. In addition, IBM model 3 also introduces *fertilities*; the fertility of a word is defined as the number of aligned words in the other language. And finally,

4. training the IBM model 4 (Brown et al., 1993) for 5 iterations; this model is identical to IBM model 3 except for the fact that it models the reordering of phrases that may be moved around as units.

Note that after obtaining the Viterbi alignments these statistical translation models are no longer used.

## 2.2 Extraction of Bilingual Phrase Pairs

Bilingual phrase pairs are automatically extracted from the word-aligned sentence pairs. Usually, the extraction of bilingual phrase pairs (Zens et al., 2002) is performed by considering all possible pairs below a certain length and ensuring that: (i) all words are consecutive, and (ii) words within the bilingual phrase pair are not aligned with words from outside.

The set $\text{BP}(w_{\text{S}1}^J, w_{\text{T}1}^I, A)$ of bilingual phrases that are extracted from the word-aligned sentence pair $(w_{\text{S}1}, \ldots, w_{\text{S}J}), (w_{\text{T}1}, \ldots, w_{\text{T}I})$ may be formally expressed as follows:

$$\text{BP}(w_{\text{S}1}^J, w_{\text{T}1}^I, A) = \{(w_{\text{S}j}^{j+m}, w_{\text{T}i}^{i+n}) :$$
$$\forall (i', j') \in A : j \leq j' \leq j + m \Leftrightarrow i \leq i' \leq i + n\}.$$

However, in our approach bilingual phrase pairs are also required to have their first and last words on both sides (source and target) aligned with at least one word in the other side.[5] Integrating these additional constraints, previous equation may be rewritten as:

$$\text{BP}(w_{\text{S}1}^J, w_{\text{T}1}^I, A) = \{(w_{\text{S}j}^{j+m}, w_{\text{T}i}^{i+n}) :$$

---

4. `http://www.fjoch.com/GIZA++.html`
5. Experiments conducted without such requirement show a significant degradation of the translation quality achieved with the inferred rules.

$$(\forall (i', j') \in A : j \leq j' \leq j + m \Leftrightarrow i \leq i' \leq i + n)$$
$$\wedge \ (\exists k \in [i, i + n] : (w_{\mathrm{S}j}, w_{\mathrm{T}k}) \in A)$$
$$\wedge \ (\exists k' \in [i, i + n] : (w_{\mathrm{S}j+m}, w_{\mathrm{T}k'}) \in A)$$
$$\wedge \ (\exists l \in [j, j + m] : (w_{\mathrm{S}l}, w_{\mathrm{T}i}) \in A)$$
$$\wedge \ (\exists l' \in [j, j + m] : (w_{\mathrm{S}l'}, w_{\mathrm{T}i+n}) \in A)\}.$$

Figure 3 shows the set bilingual phrase pairs with more than one SL word extracted from the word-aligned Spanish–English sentence pair shown in Figure 2.

## 2.3 Generalization

The generalization of the bilingual phrase pairs is simply done by using word classes instead of the words themselves; to that end, a function that maps single words into word classes is defined. The use of word classes allows the description of word reorderings, preposition changes and other divergences between SL and TL. Och and Ney (2004) use automatically obtained (Och, 1999) word classes to extract ATs for SMT. However, for RBMT, linguistically motivated word classes related to those used by the remaining modules in the MT system must be used (see section 3.1).

## 3. Alignment Templates for Shallow-Transfer Machine Translation

To apply the AT approach in a shallow-transfer MT system, the parallel corpus from which the ATs are learned must be in the intermediate representation (IR) used by the translation engine. In shallow-transfer MT the transformations to apply are mainly related to lexical forms; therefore, the IR used by the translation engine usually consists of lemma, lexical category and morphological inflection information for each word.

In order to convert the parallel corpus into the IR used by the engine, the analysis modules (morphological analyzers and part-of-speech taggers) of the engine are used to analyze both sides of the parallel corpus before computing the word alignments. After analyzing both sides of the parallel corpus we have, for each word, its lemma, lexical category and morphological inflection information. Note that generalizations are performed after word alignments and bilingual phrase pair extraction by using word classes based on that morphological information (see next section).

## 3.1 Word-Class Definition

As the transformations to apply are mainly based on the lexical category and inflection information of SL and TL words, the function that maps words into word classes will map each word into a word class representing its lexical category and morphological inflection information (such as verb, preterite tense, third person, plural).

Using the lexical category and morphological inflection information to define the set of word classes allows the method to learn general syntactic rules such as reordering and agreement rules, and verb tense changes, among others. However, in order to learn lexical changes, such as preposition changes or auxiliary verb usage, some words will be assigned single-word classes representing a lexical form, as discussed next.

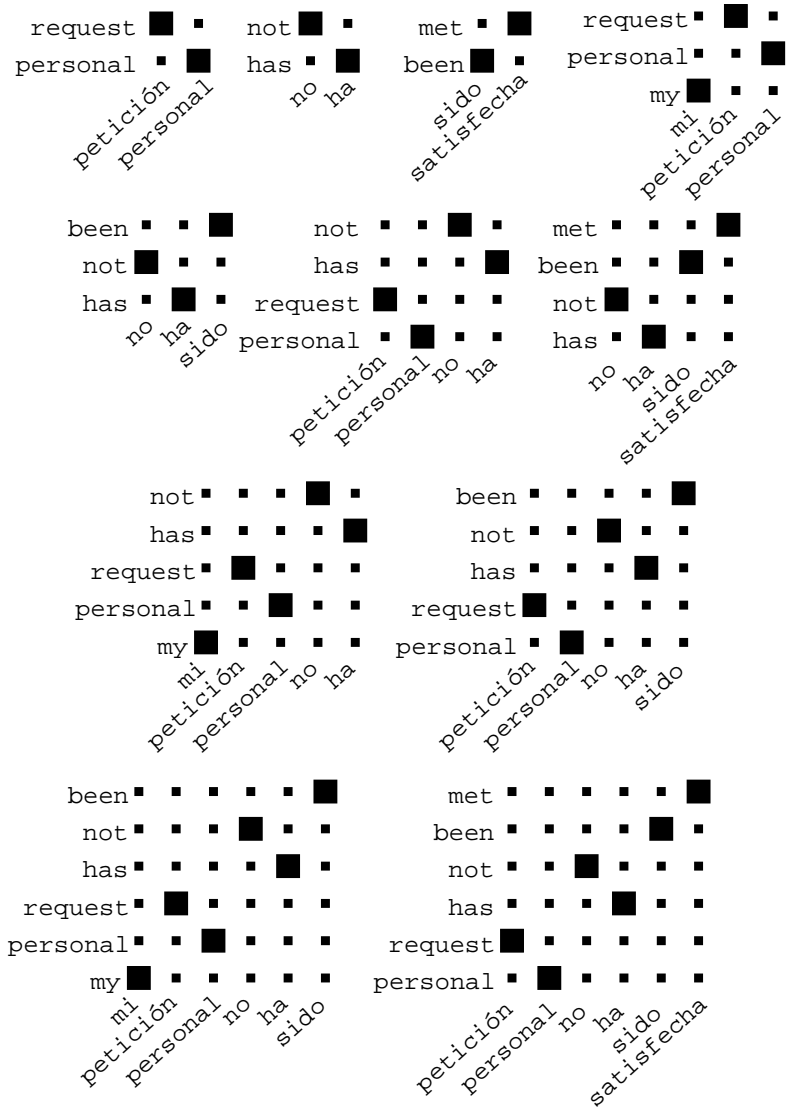

**Figure 3:** Set of bilingual phrase pairs (see section 2.2) extracted from the word-aligned Spanish–English sentence pair shown in Figure 2. Note that bilingual phrase pairs containing only one word and the whole word-aligned sentence pair have been omitted.

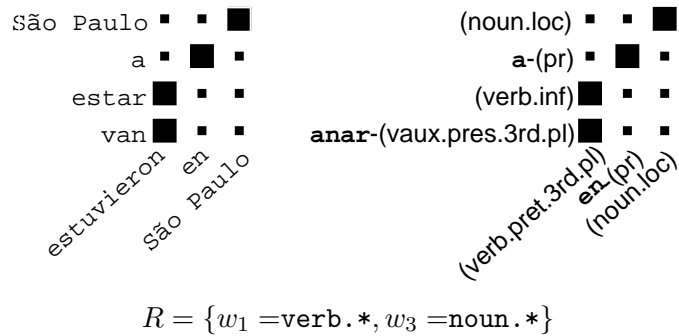

$$R = \{w_1 = \texttt{verb.*}, w_3 = \texttt{noun.*}\}$$

**Figure 4:** Example of a Spanish–Catalan bilingual phrase (left), the corresponding AT (right) obtained when each word is replaced by its corresponding word class, and TL restrictions (see section 3.2) for the Spanish-to-Catalan translation. Words in boldface correspond to lexicalized categories (see section 3.1). Word classes in the horizontal axis correspond to the SL (Spanish) and in the vertical axis to the TL (Catalan).

### 3.1.1 LEXICALIZED CATEGORIES

A set of (lexicalized) categories usually involved in lexical changes such as prepositions and auxiliary verbs may be provided. For those words whose lexical category is in the set of lexicalized categories (from now on, *lexicalized words*) the lemma is also used when defining the word class they belong to. In this way, lexicalized words are placed in single-word classes representing a particular lexical form. For example, if prepositions are considered lexicalized categories, words *to* and *for* would be in different word classes, even if they have the same lexical category and morphological inflection information, whereas words *book* and *house* would be in the same word class (noun, singular).

Typically the set of lexicalized categories is a subset of the set of closed categories, that is, those that do not grow by addition of new words to the lexicon: pronouns, auxiliary verbs, prepositions, conjunctions, etc. The most typical lexicalized words are prepositions, as they usually have many different translations depending on the SL context.

Figure 4 shows an example of a Spanish–Catalan bilingual phrase and the generalization performed when each word is replaced by its corresponding word class; words in boldface correspond to lexicalized categories. The AT shown in Figure 4 generalizes, on the one hand, the use of the auxiliary Catalan verb *anar* to express the past perfect (preterite) tense and, on the other hand, the preposition change when it refers to a location name, such as the name of a city or a country. Note that lexicalized words (e.g. `anar-(vaux.pres.3rd.pl)`, `en-(pr)`) coexist in the same AT with non-lexicalized categories (e.g. `(verb.inf)`, `(noun.loc)`) without distinction.

## 3.2 Extending the Definition of Alignment Template

In section 2 an AT was defined as a tuple $z = (S_m, T_n, A)$ in which only the alignment $A$ between SL and TL word classes was considered. Here the definition of AT is extended to $z = (S_m, T_n, A, R)$, where a set of restrictions, $R$, over the TL inflection information of the non-lexicalized categories, is added to control its application as part of a transfer rule.

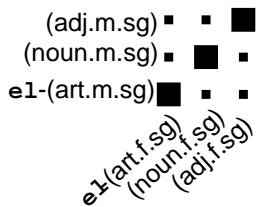

$$R = \{w_2 = \texttt{noun.m.*}, w_3 = \texttt{adj.*}\}$$

**Figure 5:** Spanish–Catalan AT and TL restrictions over the inflection information for the Spanish-to-Catalan translation (see section 3.2).

### 3.2.1 TL Restrictions

When translating (see section 4.3.1), that is, when applying the inferred ATs, the TL inflection information of non-lexicalized words is taken from the corresponding (aligned) TL word class in the AT being applied, not from the bilingual dictionary; because of this, restrictions are needed in order to prevent an AT to be applied in certain conditions that would produce an incorrect translation.

To illustrate the need for such restrictions let us consider what would happen when translating the Spanish phrase *la silla roja*[6] into Catalan by applying the extended AT shown in Figure 5, which should not be applied in this case. This AT generalizes the propagation of the masculine gender to the article and the adjective when translating a SL (Spanish) noun that is feminine singular in the SL (same with the article and the adjective) and has a masculine equivalent into Catalan, which is not the case of *silla*. After applying the extended AT in Figure 5, the *morphological generator* (see Appendix A) has to inflect the lexical form *cadira-*(`noun.m.sg`), which does not exist in Catalan,[7] as *cadira* is feminine. By taking into account some restrictions over the TL inflection information, such as the one referring to $w_2$ in the extended AT in Figure 5, we prevent the application of an AT if its application would produce an incorrect lexical form to inflect, as in the running example.

TL restrictions are obtained from the bilingual dictionary of the MT system in which the inferred transfer rules will be integrated. Bilingual dictionaries may explicitly code all the inflection information of the translation of each SL lexical form, or only the inflection information that changes from one language to the other. TL restrictions could be derived from both kinds of bilingual dictionaries; however, their extraction is easier in the second case, that is, if only changes in the inflection information are explicitly coded.

For the experiments (see section 5) the Apertium MT platform has been used; in Apertium bilingual dictionaries, only changes in inflection information are explicitly coded. The following two examples show, on the one hand, a Spanish–Catalan bilingual entry and, on the other hand, the restriction over the TL inflection information for the Spanish-to-Catalan translation derived for that bilingual entry:[8]

---

6. Translated into English as *the red chair*.

7. Note that the lexical category and morphological inflection information of the TL lexical form to inflect has been taken from the TL part of the AT.

8. Lemmas between tags `<l>` and `</l>` (left) correspond to Spanish words; analogously, lemmas between tags `<r>` and `</r>` (right) correspond to Catalan words. Lexical category and inflection information is coded through the tag `<s>` (*symbol*), the first one being the lexical category.

- Bilingual entry without any change in inflection information

  ```
  <e><p>
   <l>castigo<s n="noun"/></l>
   <r>càstig<s n="noun"/></r>
  </p></e>
  ```

  Restriction: $w$=`noun.*`

- Bilingual entry in which the gender changes from feminine (Spanish) to masculine (Catalan)

  ```
  <e><p>
   <l>calle<s n="noun"/><s n="f"/></l>
   <r>carrer<s n="noun"/><s n="m"/></r>
  </p></e>
  ```

  Restriction: $w$=`noun.m.*`

As can be seen, restrictions provide the lexical category and morphological inflection information that the lexical form should have at translation time after looking it up in the bilingual dictionary; the star at the end of each restriction means that the rest of inflection information is not restricted. The second bilingual entry would be responsible of the restrictions attached to $w_2$ in the AT shown in Figure 5. That AT can only be applied if the noun ($w_2$) is masculine in the TL (see next section to know how ATs are applied); note that the inflection information of $w_3$ is not restricted at all; this is because $w_3$ refers to an adjective that can be both masculine and feminine, as its gender depends on the gender of the noun it qualifies.

## 4. Generation of Apertium Transfer Rules

This section describes the automatic generation of Apertium structural shallow-transfer rules; note, however, that the generation of transfer rules for other shallow-transfer MT systems would also be feasible by following the approach presented here.

The structural transfer in Apertium (see appendix A) uses finite-state pattern matching to detect, in the usual left-to-right, longest-match way, fixed-length patterns of lexical forms to process and performs the corresponding transformations. A (generic) shallow-transfer rule consists of a sequence of lexical forms to detect and the transformations that need to be applied to them.

### 4.1 Discarding Useless Bilingual Phrase Pairs

Not all bilingual phrase pairs are useful in the inference of transfer rules, since the generalization that would be performed from some of them cannot be used in RBMT; more precisely, bilingual phrase pairs satisfying one or both of the following conditions are useless, and therefore, discarded:

- SL and TL non-lexicalized words are not aligned. When translating a SL non-lexicalized word (see next section) the inflection information is taken from the aligned TL word class, therefore the corresponding alignment must exist.

- the bilingual phrase pair cannot be reproduced by the MT system in which the transfer rules will be used. This happens when the translation equivalent in the bilingual dictionary differs from the one observed in the bilingual phrase. Note that TL restrictions are extracted from the bilingual dictionary, and if translation equivalents do not agree the extracted AT could end up having a set of restrictions making no sense at all.

### 4.2 Selecting the Alignment Templates to Use

To decide which ATs to take into account for the generation of rules, the method is provided with a frequency count threshold. ATs whose frequency count is below this threshold are discarded. In the experiments, two different ways of interpreting the frequency count have been tested:

- to use directly the frequency count $c$, and

- to use a modified frequency count $c' = c(1 + \log(l))$, where $l$ stands for the length of the SL part of the AT.

The second approach aims at solving the problem caused by the fact that longer ATs have lower frequency counts but may be more accurate as they take more context into account. A similar approach was used by (Mikheev, 1996) in his work on learning part-of-speech guessing rules to favor longer suffixes over shorter ones.

### 4.3 Rule Generation

A rule consists of a set $U$ of extended ATs with the same sequence of SL word classes, but different sequences of TL word classes, different alignment information or different set of TL restrictions. Formally this may be expressed as follows:

$$U = \{(S_m, T_n, A, R) \in Z : S_m = S^U\}, \tag{1}$$

where $Z$ refers to the whole set of ATs and $S^U$ to the sequence of SL word classes that all ATs in $U$ have in common. Note that each rule matches a different sequence $S^U$ of SL word classes and, therefore, there is no ambiguity in the application of the shallow-transfer rules at translation time.

Each rule $U$ is coded in Apertium's XML-based transfer language. The code generated for each rule applies always the most frequent AT in $U$ that satisfies the TL restrictions $R$; therefore, competing ATs are selected according to their frequency. A "default" AT, which translates word for word, is always added with the lowest frequency count. This AT has no TL restrictions and is the one applied when none of the remaining ATs can be applied because their TL restrictions are not met.

To check if the restrictions over the TL inflection information of an AT are met, the translation of each non-lexicalized word is retrieved from the bilingual dictionary; then, the retrieved morphological attributes (lexical category and inflection information) are compared with those specified by the corresponding restriction; the AT will be applicable if all restrictions hold.

### 4.3.1 APPLICATION OF AN ALIGNMENT TEMPLATE

The code generated in the Apertium's XML-based transfer language that applies an AT is guided by the sequence $T_n$ of TL word classes. The actions to perform for each unit in $T_n$ depend on the type of its word class:

- if the word class corresponds to a non-lexicalized word, the translation of the lemma of the aligned SL (non-lexicalized) word is retrieved by looking it up in the bilingual dictionary; then, the lexical category and morphological inflection information provided by the TL word class are attached to the translated lemma;

- if the word class corresponds to a lexicalized word, it is introduced as is; remember that word classes belonging to lexicalized words represent complete lexical forms consisting of lemma, lexical category and morphological inflection information.

Note that the information about SL lexicalized words is not taken into account when applying a given AT (just when detecting it).

The following example illustrates how the AT shown in Figure 4 would be applied in order to translate from Spanish to Catalan the input text *vivieron en Francia*.[9] This text segment, after morphological analysis and part-of-speech tagging, is transformed by the MT engine into the SL IR *vivir*-`(verb.pret.3rd.pl)` *en*-`(pr)` *Francia*-`(noun.loc)`, which becomes the input to the structural transfer module. The AT is applied in the order specified in its TL part. For the word classes corresponding to non-lexicalized words, the aligned SL words are translated into TL (Catalan) by looking them up in the bilingual dictionary: *vivir* is translated as *viure* and *Francia* is translated as *França*. Then, the inflection information provided by the TL part of the AT (see Figure 4) is attached to each translated lemma. Finally, word classes corresponding to lexicalized words are just copied to the output as they appear in the TL part of the AT. For the running example the structural transfer output would be the TL IR *anar*-`(vaux.pres.3rd.pl)` *viure*-`(verb.inf)` *a*-`(pr)` *França*-`(noun.loc)`, which the morphological generation module would transform into the Catalan phrase *van viure a França*.

---

9. Translated into English as *They lived in France.*

## 5. Experiments

The approach presented in this paper has been tested on both translation directions of the Spanish–Catalan (`es-ca`) and Spanish–Galician (`es-gl`) language pairs, and on the Spanish-to-Portuguese (`es-pt`) translation.[10,11]

The parallel corpora used for training are from different sources. The Spanish–Catalan parallel corpora come from *El Periódico de Catalunya*,[12] a daily newspaper published both in Catalan and Spanish; the Spanish–Galician parallel corpora come from *Diario Oficial de Galicia*,[13] the official publication of the autonomous government of Galicia published both in Galician and Spanish; the Spanish–Portuguese parallel corpora come from *The JRC-Acquis Multilingual Parallel Corpus* (Steinberger et al., 2006)[14] which contains European Union (EU) law applicable in the EU member states.

To test the importance of the amount of parallel corpora available for training we have used corpora of different sizes. More precisely, we have used training corpora of around 0.25, 0.5, 1.0, 1.5, and 2.0 million words in each language. The corpora were built in such a way that, for the same language pair, the larger corpora include the shorter ones. Note that word alignments have been computed from each different training corpus in isolation before the extraction of the extended ATs that are then used in the inference of shallow-transfer rules.

As was explained in section 3.1, a set of categories usually involved in lexical changes needs to be provided for the definition of word classes so as to learn not only syntactic transformations, but also lexical transformations. To that end, a small set of eight to ten lexicalized categories is used for each language. The most common lexicalized categories are: prepositions, pronouns, determiners, subordinate conjunctions, relatives, modal verbs and auxiliary verbs.

The length of the bilingual phrase pairs extracted and used to obtain the ATs has been restricted to a maximum of 7 SL words for all the experiments. Remember from section 2.2 that to extract bilingual phrases from a pair of word-aligned sentences all possible pairs (within a certain length) are considered; by restricting that length we are making the problem computationally affordable.

With respect to the frequency count threshold used to select the set of ATs to take into account (see section 4.2), we have tested frequency count thresholds between 5 and 40 for all translation tasks and AT selection criteria. The frequency count used in the evaluation is the one giving the best *translation edit rate* (TER; Snover et al., 2006) when translating a corpus, similar to the one used for testing, with 1 000 sentences (see Table 1); in Table 5 (page 627) we provide the thresholds used when the rules are inferred from the corpus with 2.0 million words in each language.

---

10. All linguistic data used can be freely downloaded from `http://sf.net/projects/apertium`, packages `apertium-es-ca-1.0.2` (around 12 800 bilingual entries), `apertium-es-gl-1.0.4` (around 10 800 bilingual entries) and `apertium-es-pt-0.9.2` (around 11 000 bilingual entries); the number of bilingual entries reported correspond to lemma-based entries.

11. A possible criticism here is that we have not used a standard translation task to test our approach; we have not done so because the Apertium linguistic resources (morphological and bilingual dictionaries) necessary for those standard tasks were not available.

12. `http://www.elperiodico.com`

13. `http://www.xunta.es/diario-oficial`

14. `http://wt.jrc.it/lt/Acquis/`

| Language pair | sentences | words |
|:---:|:---:|:---|
| `es-ca` | 1 000 | es: 22 583 |
| | | ca: 22 451 |
| `es-gl` | 1 000 | es: 22 698 |
| | | gl: 20 970 |
| `es-pt` | 1 000 | es: 23 561 |
| | | pt: 22 941 |

**Table 1:** Number of sentences and number of words in each language for the different corpora used to select the frequency count threshold used in the evaluation. The threshold finally used depends on the translation task; see Table 5 on page 627 to know which threshold has been used for each translation task when the rules are inferred from the parallel corpus with 2.0 million words in each language.

### 5.1 Evaluation

The performance of the presented approach is compared to that of the same MT system when no transfer rules are used at all (word-for-word MT), to that of the same MT system when using the hand-coded transfer rules,[15] and to that of using a state-of-the-art SMT system trained using the same parallel corpora. For the latter we have used the free/open-source SMT toolkit Moses (Koehn et al., 2007) and the SRILM language modelling toolkit (Stolcke, 2002). The training of the SMT system was done as follows:[16] First, the translation model was trained using the 90% of the training corpus. Then, a 5-gram language model was trained using the SRILM toolkit with the whole training corpus. Finally, the minimum error "rate" training algorithm (Och, 2003) used the remaining 10% of the training corpus to adjust the weight of each feature.[17] The features used by the SMT system are those used by Moses by default: 5 phrase-table features (source-to-target and target-to-source phrase translation probabilities, source-to-target and target-to-source lexical weightings, and phrase penalty), a distance-based cost (total number of word movements), the sentence word count, and the TL model.

Translation performance is evaluated using two different measures; on the one hand, the *translation edit rate* (TER; Snover et al., 2006), and on the other hand, the bilingual evaluation understudy (BLEU; Papineni et al., 2002); in both cases the same evaluation corpora have been used and the confidence intervals of the measures being reported are given (see below).

#### 5.1.1 Confidence intervals

Confidence intervals of MT quality measures are calculated through the *bootstrap resampling* method as described by Koehn (2004). In general, the bootstrap resampling method consists of estimating the precision of sample statistics (in our case, translation quality measures) by randomly resampling with replacement (that is, allowing repetitions) from the full set of samples (Efron & Tibshirani, 1994); in MT, sentences and their respective reference translations. This method has the property that no assumptions are made about the underlying distribution of the variable, in our case, the MT quality measure.

---

15. Those in the corresponding Apertium language packages.
16. For detailed training instructions visit `http://www.statmt.org/wmt09/baseline.html`.
17. The minimum error "rate" training used BLEU as an evaluation measure.

| Language pair | sentences | words |
|:---:|:---:|:---|
| `es-ca` | 2 400 | es: 55 064 |
| | | ca: 54 730 |
| `es-gl` | 2 450 | es: 55 826 |
| | | gl: 51 603 |
| `es-pt` | 2 000 | es: 55 814 |
| | | pt: 53 762 |

**Table 2:** Number of sentences and number of words in each language for the different test corpora used for evaluation.

The calculation of the confidence intervals consists of the following steps:

1. the translation performance is evaluated a large number of times, in our experiments 1 000 times, using randomly chosen sentences from the test corpus, and their counterpart sentences in the reference corpus;

2. all the calculated measures are sorted in ascending order; and

3. the top $q\%$ and the bottom $q\%$ elements are removed from that list.

After that, the remaining values are in the interval $[a, b]$. This interval approximates with probability $1 - 2q/100$ the range of values in which the quality measure being reported lies for test corpora with a number of sentences equal to that used to carry out the evaluation.

### 5.1.2 Evaluation corpora

Table 2 shows the number of sentences and the number of SL and TL words of the different test corpora used for the evaluation of the inferred rules for each translation being considered. These test corpora come from independent parallel corpora, from a different source, with no relation to those used for training. More precisely, the test corpora for Spanish–Catalan and Spanish–Galician comes from *Revista Consumer Eroski* (Alcázar, 2005),[18] a magazine addressed to consumers published in Spanish, Catalan, Galician and Basque; the test corpora for Spanish–Portuguese comes from the shared evaluation task of the 2008 workshop on SMT.[19]

### 5.2 Results

Figure 6 shows the TER and BLEU scores, together with their respective 95% confidence intervals, achieved for each translation direction of the Spanish–Catalan language pair when using training corpora of different sizes. The error rates reported are: (a) the results when the frequency count is directly used to select the set of ATs to use for the rules generation, (b) the results achieved by a state-of-the-art SMT system trained on the same corpora, (c) the results achieved when using hand-coded transfer rules, and (d) the results of a word-for-word translation (when no structural transformations are applied). The results achieved when the modified frequency count described in section 4.2 is used to select the set of ATs to use are not reported since they are indistinguishable in practice from those achieved by using directly the frequency count; for this reason, they will not be considered in the rest of the experiments. Notice that in all cases, except for the SMT results, the same linguistic data (morphological and bilingual dictionaries) have been used. Some Catalan-to-Spanish translations produced by the automatically inferred rules are shown in Table 3.

---

18. `http://revista.consumer.es`
19. `http://www.statmt.org/wmt08/wmt08-eval.tar.gz`

| Catalan (SL) | Spanish (TL) |
|---|---|
| ... els gossos catalogats de perillosos han de tenir una assegurança ... | ... los perros catalogados de peligrosos **deben** $\phi$ tener seguro ... |
| ... es va descobrir en el cacauet ... | .. se $\phi$ **descubrió** en el cacahuete ... |
| ... va tenir un infart de miocardi ... | ... $\phi$ **tuvo** un infarto de miocardio ... |
| ... els fonaments científics per considerar funcionals diversos aliments són ... | los fundamentos científicos **para** considerar funcionales varios alimentos son ... |
| ... l'enveja no es manifesta ... | **la** envidia no se manifiesta ... |
| ... cal preservar-lo de la llum ... | ... **hay que** preservarlos de la luz ... |

**Table 3:** Translation examples for the Catalan-to-Spanish translation. The translations reported are those produced when using the automatically inferred rules; words in boldface indicate changes with respect to a word-for-word translation; $\phi$ indicates a word deleted with respect to a word-for-word translation.

Results in Figure 6 show that, as expected, the translation quality achieved by the inferred transfer rules is better than that of a word-for-word translation, even when a small parallel corpus with around 0.5 million words in each language is used; note however, that in the case of the Spanish-to-Catalan translation confidence intervals overlap for a training corpus of 0.25, 0.5 and 1.0 million words, the overlap being smaller for the latter.

Results in Figure 6 also show that a SMT system performs worse than the rules automatically inferred from the same parallel corpus and even worse than a word-for-word translation. This is because the training corpora we have used are not large enough to learn a wide-coverage bilingual lexicon and, consequently, most of the words to translate are unknown to the SMT system. Remember that our approach only learns transfer rules from the parallel corpus, not bilingual entries, and that the same bilingual dictionary is used by the hand-coded rules, the automatically inferred rules and the word-for-word translation. In section 5.2.1 (page 626) we discuss the results achieved by the SMT system when the bilingual dictionary in the corresponding Apertium package is added to the SMT training data.

Figure 7 shows, for each translation direction of the Spanish–Galician language pair, the same MT quality measures and for the same translation setups reported for Spanish–Catalan in Figure 6.

The Spanish–Galician language pair shows results in agreement to those obtained for Spanish–Catalan; however, the improvement on the Galician-to-Spanish translation quality, compared to word-for-word translation, is smaller. In addition, the improvement obtained in the case of Spanish–Catalan by increasing the amount of corpora used for training is greater than that of Spanish–Galician, as shown by the slope of the curve. This is because the significant frequent patterns which can be learned from the training corpora are selected very early. Note that our method is unlikely to perform worse than word-for-word translation (when no rules are used).

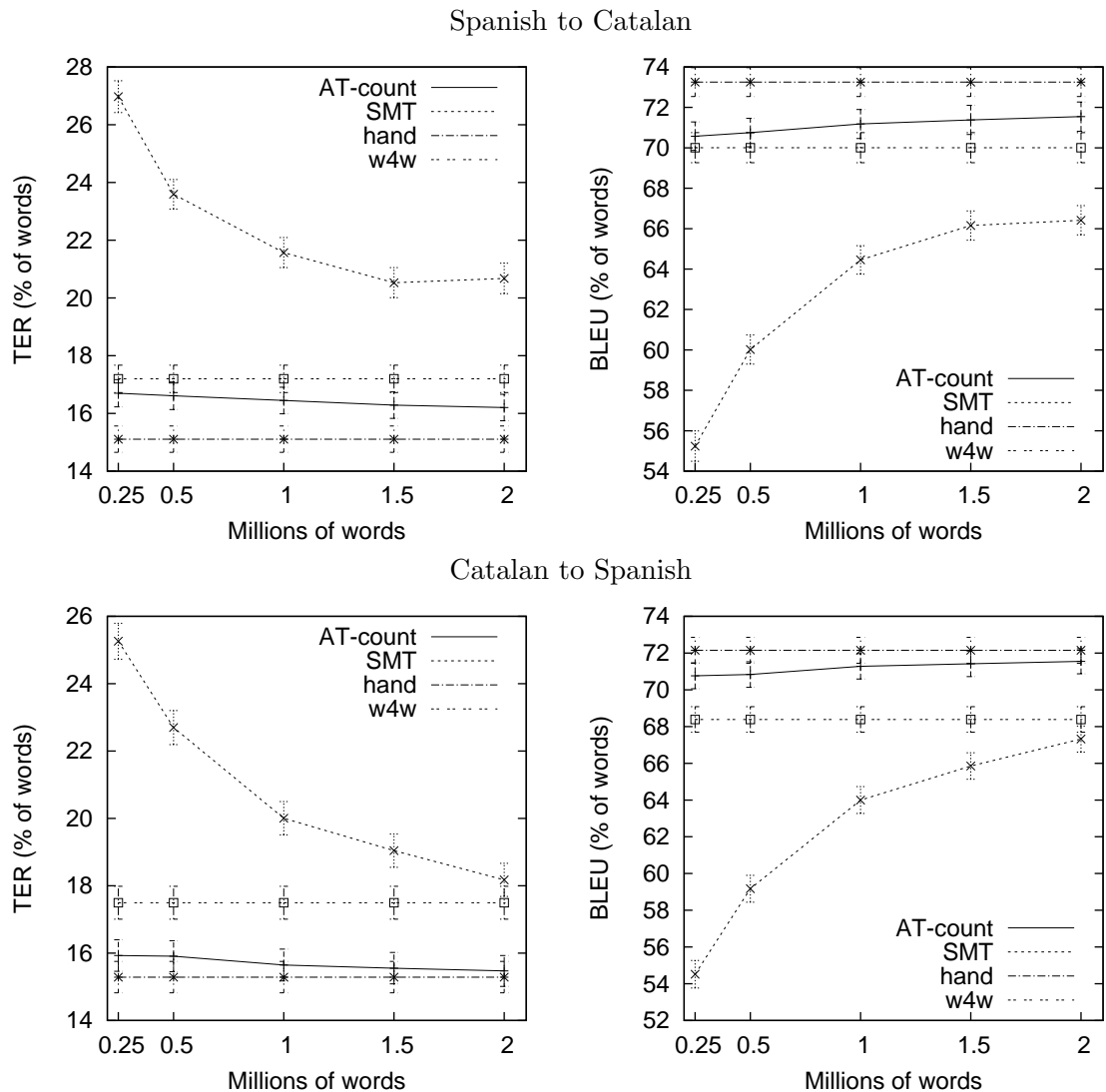

**Figure 6:** TER and BLEU scores (vertical axis), with their respective 95% confidence intervals, for each translation direction of the Spanish–Catalan language pair when using training corpora of different sizes (horizontal axis). AT-count refers to the result achieved when the count is directly used to select the set of ATs to use; SMT refers to the result achieved by a state-of-the-art SMT system trained on the same parallel corpora; hand refers to the results achieved when hand-coded transfer rules are used; w4w ("word for word") refers to the result achieved when no transfer rules are used.

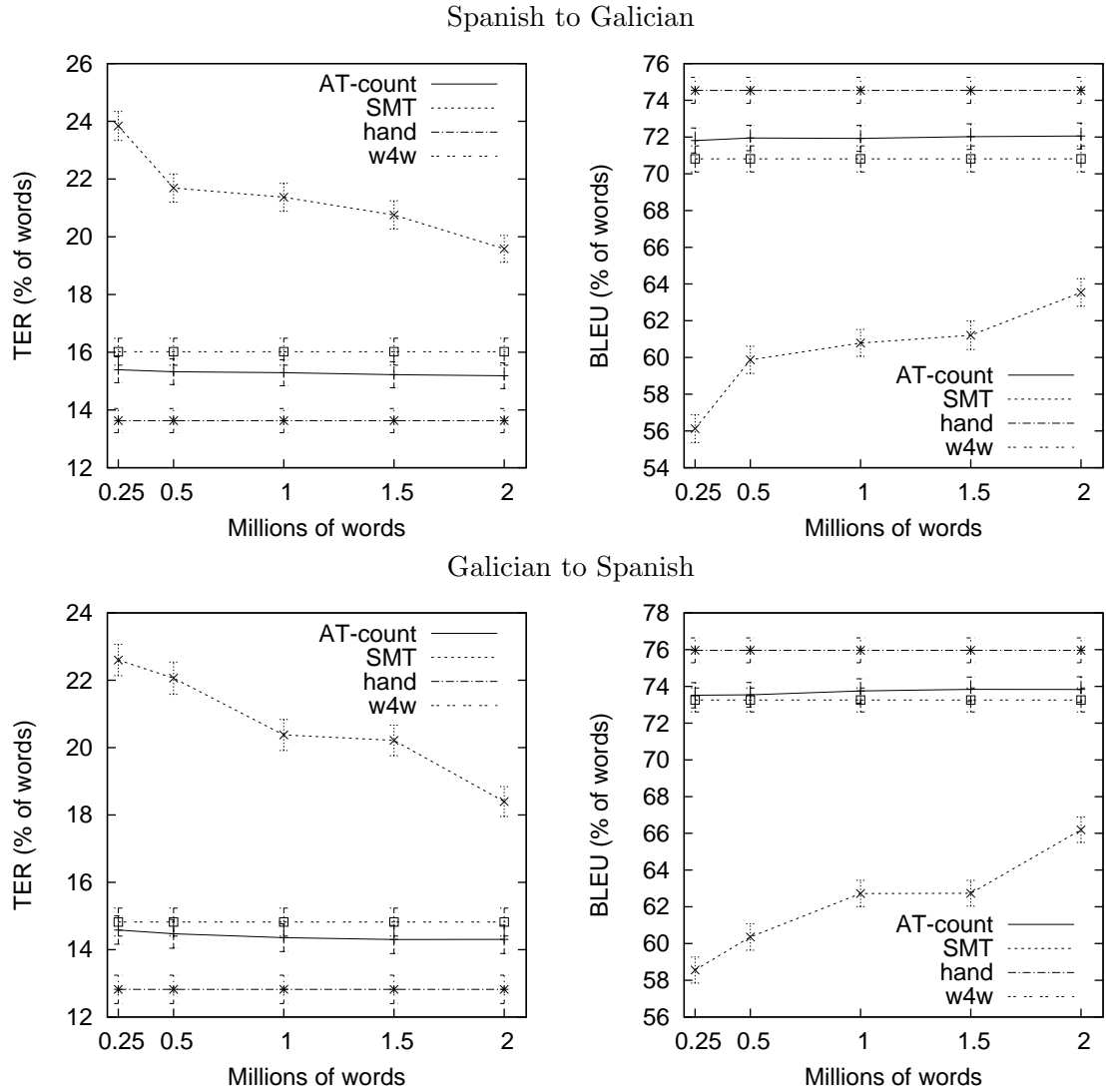

**Figure 7:** TERs and BLEU scores (vertical axis), with their respective 95% confidence interval, for each translation direction of the Spanish–Galician language pair when using training corpora of different sizes (horizontal axis). The measures reported correspond to the results achieved when using different MT setups (as described in Figure 6).

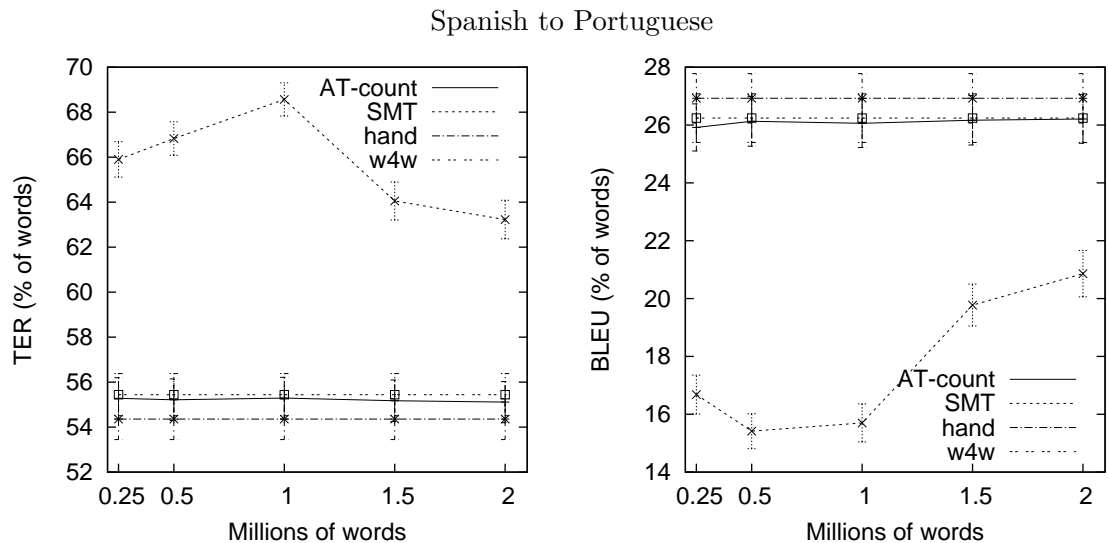

**Figure 8:** TER and BLEU scores (vertical axis), with their respective 95% confidence intervals, for the Spanish–to-Portuguese translation when using training corpora of different sizes (horizontal axis). The measures reported correspond to the results achieved when using different MT setups (see Figure 6).

Concerning the Spanish-to-Portuguese translation, Figure 8 shows the TER and BLEU scores achieved for the different sizes of training corpora used. Notice that the automatically inferred rules perform better than the word-for-word translation, although their confidence intervals show a large overlap. It is worth mentioning that the confidence intervals obtained for the hand-coded transfer rules also overlap with those of the automatically inferred rules and the word-for-word translation. As in the rest of experiments the SMT system performs worse because the training corpus is not large enough to learn a wide-coverage bilingual lexicon.

The difference between the results achieved when using hand-coded transfer rules and that of using no rules at all (word-for-word translation) is very small compared to the rest of translation tasks considered in this paper. Moreover, the TER and BLEU scores obtained are very poor although Spanish and Portuguese are two related languages, and, therefore, translating between them should not be such a difficult task. Indeed, an evaluation of the hand-coded transfer rules performed using an evaluation corpus in which the reference translation is a post-edited (corrected) version of the MT output produced with the same hand-coded rules shows a TER below 10%.

The poor results obtained for the Spanish-to-Portuguese may be explained by the fact that the evaluation corpus, as well as the training corpora used, may have not been built by translating one language (say Spanish or Portuguese) into the other, but by translating from a third language (possibly English or French).[20] This causes the reference translation to be very different compared to the translations automatically performed, thus giving very high TERs. On the other hand, this may also cause the alignments obtained from the training corpora to be unreliable, as shown by the percentage of discarded bilingual phrase pairs. This percentage is, for all training corpora, around 54% for the Spanish-to-Portuguese translation, about 22% for the Spanish–Catalan language pairs, and around 20% for the Spanish–Galician language pair.

| | AT-count | | SMT+dictionary | | SMT | |
|---|---|---|---|---|---|---|
| | **TER** | **OOV (%)** | **TER** | **OOV (%)** | **TER** | **OOV (%)** |
| es-ca | [15.8, 16.7] | 4.3% | [18.0, 19.0] | 3.4% | [20.1, 21.2] | 5.7% |
| ca-es | [15.0, 15.9] | 4.9% | [15.5, 16.4] | 3.8% | [17.7, 18.7] | 6.1% |
| es-gl | [14.7, 15.6] | 9.3% | [16.2, 17.1] | 6.9% | [19.1, 20.0] | 18.7% |
| gl-es | [13.9, 14.7] | 10.2% | [13.6, 14.4] | 8.2% | [18.0, 18.8] | 21.1% |
| es-pt | [54.2, 56.0] | 3.8% | [57.2, 59.0] | 3.1% | [62.4, 64.1] | 12.6% |

**Table 4:** 95% confidence intervals for the TER and ratio of out-of-vocabulary (OOV) words when the test corpus is translated: with the rules automatically obtained from parallel corpora (AT-count), with a SMT system trained with the same parallel corpora (SMT), and with a SMT system trained on the same parallel corpora plus the corresponding Apertium bilingual dictionary (SMT+dictionary). The data reported correspond to the case in which the training corpus has 2.0 million words in each language.

### 5.2.1 Adding the bilingual dictionary to the SMT training data

With the aim of testing whether the difference in the translation performance between the shallow-transfer rules and the SMT system is due to the fact that Apertium uses a wide-coverage, manually-built bilingual dictionary, we have added the bilingual dictionary in the corresponding Apertium package to the SMT training data (Tyers et al., 2009).[21] It is worth noting that adding the bilingual dictionary to the training corpus does not only improve the vocabulary coverage of the SMT systems inferred, but also helps the word alignment process by adding word-to-word alignment, which gives an additional advantage to the SMT system with respect other systems; the bilingual dictionary has not been added to corpus used to learn the AT used for the automatic inference of shallow-transfer rules.

Table 4 shows the 95% confidence intervals for the TER and the ratio of out-of-vocabulary (OOV) words when the test corpus is translated by means of Apertium with the shallow-transfer rules automatically obtained form the parallel corpus with 2.0 million words in each language (AT-count); when it is translated using a SMT system trained on this same parallel corpus (SMT); and, when it is translated with a SMT system trained with a par-allel corpus containing the original corpus with 2.0 million words in each language plus the corresponding Apertium bilingual dictionary (SMT+dictionary).

---

20. Remember that the training corpora contains European Union law and that the evaluation corpus comes from the European Parliament proceedings.

21. Apertium bilingual dictionaries contain lemma-based bilingual entries which have been expanded to include all possible inflected forms before adding them to the SMT training data. After inflecting all the lemma-based bilingual entries the bilingual dictionary added to the SMT training data consists of (approximately) 1.8 million entries for Spanish–Catalan, 1.2 million entries for Spanish–Galician, and 0.9 million entries for Spanish–Portuguese.

|  | freq. count | number of rules | rules used | % used | % performing word-for-word |
|---|---|---|---|---|---|
| `es-ca` | 8 | 32 165 | 8 133 | 25.3% | 2.77% |
| `ca-es` | 14 | 17 930 | 6 785 | 37.8% | 2.08% |
| `es-gl` | 13 | 14 764 | 3 777 | 25.6% | 1.16% |
| `gl-es` | 6 | 28 573 | 4 898 | 17.1% | 1.51% |
| `es-pt` | 25 | 5 402 | 2 636 | 48.8% | 1.18% |

**Table 5:** For each translation task, the following data are shown: the frequency count threshold used, the number of rules generated, the number (and percentage of rules) that are used in the translation of the corresponding evaluation corpus, and the percentage of rule applications that end up performing a word-for-word translation. The data reported correspond to the rules obtained from the training corpora with 2.0 million words in each language.

The results in Table 4 show that, as expected, the SMT results improve when the bilingual dictionary is added to the training corpus; note however, that the results obtained for `es-ca`, `ca-es`, `es-gl`, and `es-pt` are still worse than those achieved by the automatically inferred rules, although the `ca-es` "SMT+dictionary" confidence interval shows a large overlap with that of the automatically inferred rules. The only translation task in which the "SMT+dictionary" system provides better results than the automatically inferred rules is the `gl-es` task, although its confidence interval overlaps with that of the automatically inferred rules. In all cases the ratio of OOV words for "SMT+dictionary" is below that of the automatically inferred rules because some words not present in the bilingual dictionary do appear in the training corpus.

### 5.2.2 Analysis of the inferred rules

Table 5 shows, for each translation task, the frequency count threshold used in the generation of rules, the number of rules obtained and the number of them that are used in the translation of the corresponding evaluation corpus; remember that the frequency count threshold used in each translation task is the one minimizing the TER when translating the corpora described in Table 1. The data reported in Table 5 correspond to the rules inferred with the largest training corpora (2.0 million word in each language). Note that the number of inferred rules varies depending on the translation task; for instance, the number of rules for `es-ca` is around twice the number of rules for `ca-es`, this is because to produce the minimum TER less rules happen to be needed in the case of `ca-es`.

The data in Table 5 reveal, on the one hand, that the percentage of rules finally used to translate the corresponding evaluation corpus varies depending on the translation task, and, on the other hand, that the percentage of rules that end up applying the "default" AT (which performs a word-for-word translation, see section 4) depends on the translation task, although it is always below 3%.

Figure 9 shows for the Spanish-to-Catalan translation, on top, the number of rules obtained and the number of rules used in the translation of the evaluation corpus, all grouped by rule length (number of SL word classes); and, at the bottom, the number of rule applications and the number of rule applications that end up performing a word-for-word translation (apply the "default" AT); in the generation of the rules a frequency count threshold of 8 was used. Notice that there are rules of unit length, i.e. rules that process only a single SL word class: they are needed because the bilingual dictionary leaves some translation decisions open, such as the gender and number of some words that can be both masculine and feminine, or singular and plural, in the TL. The data in that figure correspond to the rules inferred form the largest training corpora used; in any case, with the rest of training corpora, a similar behaviour is obtained; the same happens with the remaining translation tasks.

Figure 9 shows that most of the rules generated process SL patterns of only 3 or 4 word classes; the number of rules processing 7 SL word classes being very low. Remember that for the extraction of bilingual phrase pairs their length was restricted to 7 SL words.

Finally, it is worth mentioning that the number of inferred rules is very high compared to the number of hand-coded rules. Note, however, that automatically inferred rules are more specific and lexicalized than hand-coded ones. Hand-coded rules use macros and complex control flow statements which allow them to treat more phenomena in the same rule.

## 6. Discussion

This paper has focused on the inference of structural transfer rules to be used in MT, and more precisely on the inference of shallow-transfer rules. It describes how to extend the AT approach introduced in the SMT framework in order to use it to generate shallow-transfer rules to be used in RBMT. To this end, a very small amount of linguistic information, in addition to the linguistic data used by the MT engine, has been used in order to learn not only syntactic changes, but also lexical changes to apply when translating SL texts into TL. This linguistic information consists of a small set of lexical categories involved in lexical changes (prepositions, pronouns, etc.) and can easily be provided by an expert.

The approach has been tested using data from three existing language pairs of the free/open-source shallow-transfer MT engine Apertium; more precisely, the presented approach has been tested on both translation directions of the Spanish–Catalan and Spanish–Galician languages pairs, and on the Spanish-to-Portuguese translation. For each language pair, training corpora of different sizes have been used so as to test the importance of the size of training corpora available.

The evaluation has been done, in all cases, by using independent parallel corpora, coming from an independent source, with no relation with the parallel corpora used for training. In the evaluation the translation quality achieved by the automatically inferred rules has been compared to that of using hand-coded shallow-transfer rules, to that of a word-for-word translation, and to that of using a state-of-the-art SMT system trained on the same parallel corpora. In all cases the automatically inferred rules perform better than the SMT system; moreover, when the Apertium bilingual dictionary is added to the SMT training data only one translation task performed slightly better than the automatically inferred rules. Notice that our approach, unlike that by Caseli et al. (2006), is aimed at learning shallow-transfer rules, not bilingual entries, and that we have used the bilingual dictionary provided by the corresponding Apertium language-pair package.

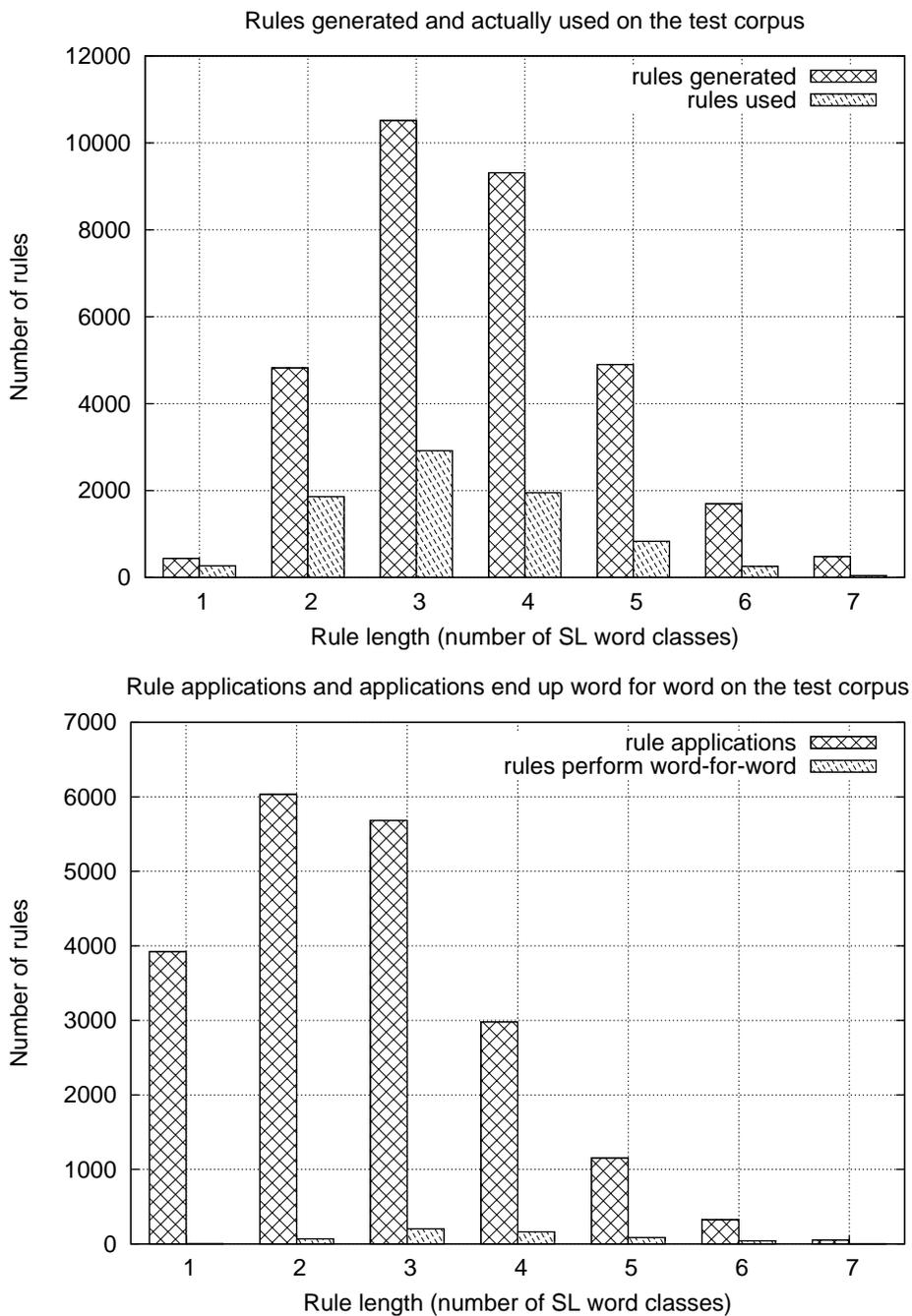

**Figure 9:** For the Spanish-to-Catalan translation, rules generated and used in the translation of the corresponding evaluation corpus (top), and number of rule applications and number of those applications that end up performing a word-for-word translation (bottom). Reported data are grouped by rule length (number of SL word classes).

The evaluation of the inferred rules for both translation directions of the Spanish–Catalan and the Spanish–Galician language pairs show an improvement in the translation quality as compared to word-for-word translation, even when a very small parallel corpus is used. In the case of the Spanish-to-Portuguese translation, there is a very small improvement: confidence intervals show a large overlap.

To our knowledge, this is the first time that the AT approach is extended for its use in RBMT; an important property of the inferred rules is that they can be edited by human experts so as to improve them. This means that developers of RBMT systems can use this method to obtain a set of initial transfer rules that can be then refined by linguists; proceeding in this way, human experts can focus on the more difficult issues of writing accurate transfer rules for MT, as most of the required rules are automatically obtained from parallel corpora. From our point of view, this is a great advantage over other corpus-based approaches to MT, such as SMT, because, in our approach, automatically generated rules can coexist with hand-coded ones.

With respect to the parallel corpus used for training, the results achieved by the inferred rules for the Spanish-to-Portuguese translation show that the procedure followed to build the parallel corpus, that is, the way in which the translation from one language into the other one is performed, deserves special attention. In our opinion, it may be concluded that parallel corpora that have been built by translating from a third language may not be appropriate for the task of inferring rules to be used in RBMT, especially if the languages involved are closely related and the third language is not.

It must be mentioned that software implementing the method described in this paper has been released as free/open-source software under the GNU GPL license[22] and can be freely downloaded from `http://apertium.sf.net`, package name `apertium-transfer-tools`. The public availability of the source code ensures the reproducibility of all the experiments conducted and allows other researchers to improve the approach discussed here, saving them from having to implement the algorithms all over again. In addition, the method has been implemented in such a way that it integrates with the Apertium free/open-source MT platform (see appendix A); this benefits, on the one hand, other research that uses Apertium as a research platform, and on the other hand, people developing new language pairs for Apertium.

We plan to improve the generated rules by using linguistic criteria for the extraction of the bilingual phrase pairs that are generalized to ATs. Note that in the experiments reported in this paper bilingual phrase pairs are extracted from the training corpus without worrying whether they are well-formed syntactic constituents or not. We also plan to study how to use lexicalized categories in a more flexible way. It would be of interest to have context-dependent lexicalized categories, that is, categories which are lexicalized only in some contexts, while not in others; this would improve the generalization performed by the extended ATs and reduce the number of inferred rules.

Another improvement we plan to achieve is the extension of the present approach so that rules for translation between less-related language pairs can be inferred. Recently, the transfer in Apertium has been extended to translate between more divergent languages by splitting the structural transference phase into 3 stages: the first one detects word patterns called chunks; the second one operates with sequences of chunks; finally, the third one makes some "finishing" operations within the chunks detected in the first stage. Our approach could be extended by detecting chunks in the training parallel corpus using linguistic criteria as mentioned in the previous paragraph, or using the "Marker Hypothesis" (Green, 1979), as done by Gough and Way (2004), and then extracting ATs based on chunk classes instead of word classes, as it is done now. In any case, it would be worth testing the method in its present for the translation between less-related languages by using longer ATs and larger training corpora.

---

22. `http://www.gnu.org/licenses/gpl-2.0.html`

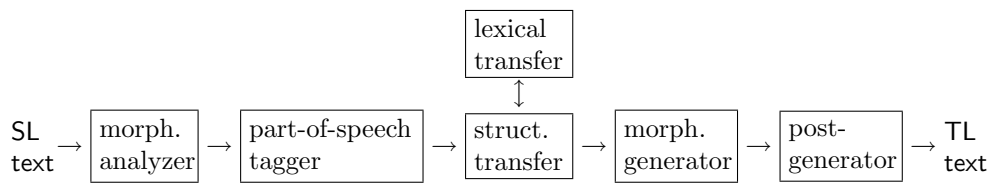

**Figure 10:** Main modules of the free/open-source shallow-transfer MT engine Apertium used in the experiments (see appendix A).

## Acknowledgments

Work funded by the Spanish Ministry of Education and Science and the European Social Fund through research grant BES-2004-4711, by the Spanish Ministry of Industry, Tourism and Commerce through projects TIC2003-08681-C02-01, FIT340101-2004-3 and FIT-350401-2006-5, and by the Spanish Ministry of Education and Science through project TIN2006-15071-C03-01. The authors thank the anonymous referees for suggesting significant improvements to this paper and Francis Tyers for proof-reading it.

## Appendix A. The Apertium Machine Translation Platform

This appendix briefly describes the free/open-source shallow-transfer MT engine Apertium[23] (Armentano-Oller et al., 2006) used for the experiments. Apertium follows the shallow-transfer approach shown in Figure 10:

- A *morphological analyzer* which tokenizes the text in surface forms and delivers, for each surface form, one or more *lexical forms* consisting of *lemma*, *lexical category* and morphological inflection information.

- A *part-of-speech tagger* (categorial disambiguator) which chooses, using a first-order hidden Markov model (Cutting et al., 1992; Baum & Petrie, 1966), one of the lexical forms corresponding to an ambiguous surface form.

- A *lexical transfer* module which reads each SL lexical form and delivers the corresponding TL lexical form by looking it up in a bilingual dictionary.

---

23. The MT engine, documentation, and linguistic data for different language pairs can be downloaded from `http://apertium.sf.net`.

- A *structural transfer* module (parallel to the lexical transfer) which uses a finite-state chunker to detect patterns, such as "article–noun–adjective", of lexical forms which need to be processed for word reorderings, agreement, etc., and then performs these operations. This is the module that applies the structural transfer rules automatically inferred from parallel corpora using the method in this paper.

- A *morphological generator* which delivers a TL surface form for each TL lexical form, by suitably inflecting it.

- A *post-generator* which performs orthographic operations such as contractions (e.g. Spanish *de+el → del*) and apostrophations (e.g. Catalan *el+institut → l'institut*).

The Apertium MT engine is completely independent from the linguistic data used to translate for a given language pair. Linguistic data is coded using XML-based formats,[24] which allows for easy data transformation and maintenance.

---

24. The XML formats (http://www.w3.org/XML/) for each type of linguistic data are defined through conveniently-designed XML document-type definitions (DTDs) which may be found inside the Apertium package.